\documentclass[letterpaper, 10 pt, journal, twoside]{IEEEtran}
\IEEEoverridecommandlockouts

\usepackage[bottom]{footmisc}
\usepackage{blindtext}
\usepackage[utf8]{inputenc}
\usepackage{amsmath}
\usepackage{amssymb}
\usepackage{color}
\usepackage{graphicx}
\usepackage[pdfencoding=auto]{hyperref}
\usepackage{bm}
\usepackage{graphicx}
\usepackage{comment}
\usepackage{soul, xcolor}
\usepackage{pdfpages}
\usepackage{subfigure}
\usepackage{tikz}
\usepackage{float}
\usepackage{perpage} 
\MakePerPage{footnote}
\usepackage{pgfplots, alphalph}
\usepackage{sansmath,tkz-base}
\usepackage{multicol}
\pgfplotsset{
	tick label style = {font=\sffamily\footnotesize},
	every axis label/.append style={font=\sffamily\footnotesize},
	legend style={font=\tiny\footnotesize},
}
\sethlcolor{green}
 	
\usepackage{booktabs}
\usepackage{arydshln}

\newcommand{\diag}{\mathrm{diag}}

\newcommand{\norm}[1]{\left\lVert#1\right\rVert}

\newcommand{\bfalpha}{\boldsymbol{\alpha}}
\newcommand{\bfbeta}{\boldsymbol{\beta}}

\newcommand{\bfmu}{\boldsymbol{\mu}}

\newcommand{\bfxi}{\boldsymbol{\xi}}

\newcommand{\bfrho}{\boldsymbol{\rho}}
\newcommand{\bftau}{\boldsymbol{\tau}}

\newcommand{\bfUpsilon}{\boldsymbol{\Upsilon}}

\newcommand{\bfPsi}{\boldsymbol{\Psi}}

\newcommand{\bfa}{\ensuremath {\bm{a}}}
\newcommand{\bfb}{\ensuremath {\bm{b}}}
\newcommand{\bfc}{\ensuremath {\bm{c}}}

\newcommand{\bff}{\ensuremath {\bm{f}}}

\newcommand{\bfh}{\ensuremath {\bm{h}}}

\newcommand{\bfp}{\ensuremath {\bm{p}}}
\newcommand{\bfq}{\ensuremath {\bm{q}}}

\newcommand{\bfv}{\ensuremath {\bm{v}}}
\newcommand{\bfw}{\ensuremath {\bm{w}}}
\newcommand{\bfx}{\ensuremath {\bm{x}}}

\newcommand{\bfA}{\mathbf{A}}

\newcommand{\bfC}{\mathbf{C}}

\newcommand{\bfG}{\mathbf{G}}

\newcommand{\bfP}{\mathbf{P}}

\newcommand{\bfR}{\mathbf{R}}

\newcommand{\bfU}{\mathbf{U}}

\newcommand{\bfX}{\mathbf{X}}
\newcommand{\bfY}{\mathbf{Y}}

\newcommand{\calK}{{\cal K}}

\newcommand{\calQ}{{\cal Q}}

\newcommand{\calW}{{\cal W}}

\newcommand{\bbR}{{\mathbb{R}}}

\title{Humanoid Control Under Interchangeable Fixed and Sliding Unilateral Contacts}
\author{Saeid Samadi, Julien Roux, Arnaud Tanguy, St\'{e}phane Caron, Abderrahmane Kheddar
	\thanks{S.~Samadi, J.~Roux and A.~Kheddar are with Laboratory of Informatics,
		Robotics and Microelectronics (LIRMM), CNRS-University of Montpellier, France.
		Corresponding author: {\tt\footnotesize saeid.samadi@lirmm.fr}}
	\thanks{A.~Kheddar and A. Tanguy are with CNRS-AIST Joint Robotics Laboratory (JRL), IRL, Tsukuba, Japan.}
	 \thanks{S.~Caron is with ANYbotics AG, Zürich, Switzerland. This work was carried while he was with CNRS, France.}%
}

\begin{document}
\maketitle

\begin{abstract}
In this letter, we propose a whole-body control strategy for humanoid robots in multi-contact settings that enables switching between fixed and sliding contacts under active balance. 
We compute, in real-time, a safe center-of-mass position and wrench distribution of the contact points based on the Chebyshev center. Our solution is formulated as a quadratic programming problem without \emph{a priori} computation of balance regions.
We assess our approach with experiments highlighting switches between fixed and sliding contact modes in multi-contact configurations. A humanoid robot demonstrates such contact interchanges from fully-fixed to multi-sliding and also shuffling of the foot.
The scenarios illustrate the performance of our control scheme in achieving the desired forces, CoM position attractor, and planned trajectories while actively maintaining balance.
\end{abstract}


\section{Introduction} 
\label{Sec_Introduction}

\IEEEPARstart{M}{ulti-contact} motion is a key technology to increase humanoid robots' loco-manipulation abilities in confined and narrow spaces~\cite{kheddar2019ram}. Yet, state-of-the-art multi-contact planning and control considers only creating and breaking contacts to support the motion~\cite{bouyarmane2018hhb}. In many situations, however, switching contacts through releasing one of the established ones is not possible. This is the case, for example, in narrow or cumbersome spaces where free space is limited. Another example is when balance cannot be kept by breaking any of the existing contacts. In such cases, using sliding contacts to support the motion and the balance is an alternative. There are other contexts where tasks require sliding to be controlled (e.g. sanding or surface smoothing). In this letter, we address whole-body humanoid multi-contact task-space control allowing interchangeable multi-contact transitions between fixed (creating and removing) and sliding ones.     

Current researches and control strategies for humanoid robots deal with multi-contact motions. Complex multi-contact motions are found in non-gaited or acyclic locomotion~\cite{reher2020springer,kumagai2019humanoids}, ladder climbing~\cite{nozawa2016humanoids,vaillant2016springer}, grasping~\cite{collette2008icra}, manipulation~\cite{garcia-haro2019humanoids}, impact generations~\cite{wang-dehio2020impactaware}... to cite a few. Such skills shall be achieved under active balance of the humanoid robot~\cite{kajita2010iros,caron2019icra,morisawa2018humanoids,morisawa2019iros}.

To enforce dynamic balance for multi-legged robots in multi-contact; recent methods suggest computing gravito-inertial wrench cones (GIWC)~\cite{caron2015rss}, Zero-tilting moment points (ZMP)~\cite{caron2017tro,harada2006tro} and center-of-mass (CoM) support polygons~\cite{audren2018tro}, see also~\cite{bouyarmane2018hhb}. These methods end-up computing regions (e.g. polytopes, polyhedral cones...) where the CoM position or the wrenches should live. Reducing the cost of such computations is the main challenge to integrate them in control and reactive planning. Indeed, they are mostly used in planning. We propose a decent alternative with a lighter solution to tackle this shortcoming.
\begin{figure}[!t] 
	\centering
	\includegraphics[height=.75\columnwidth]{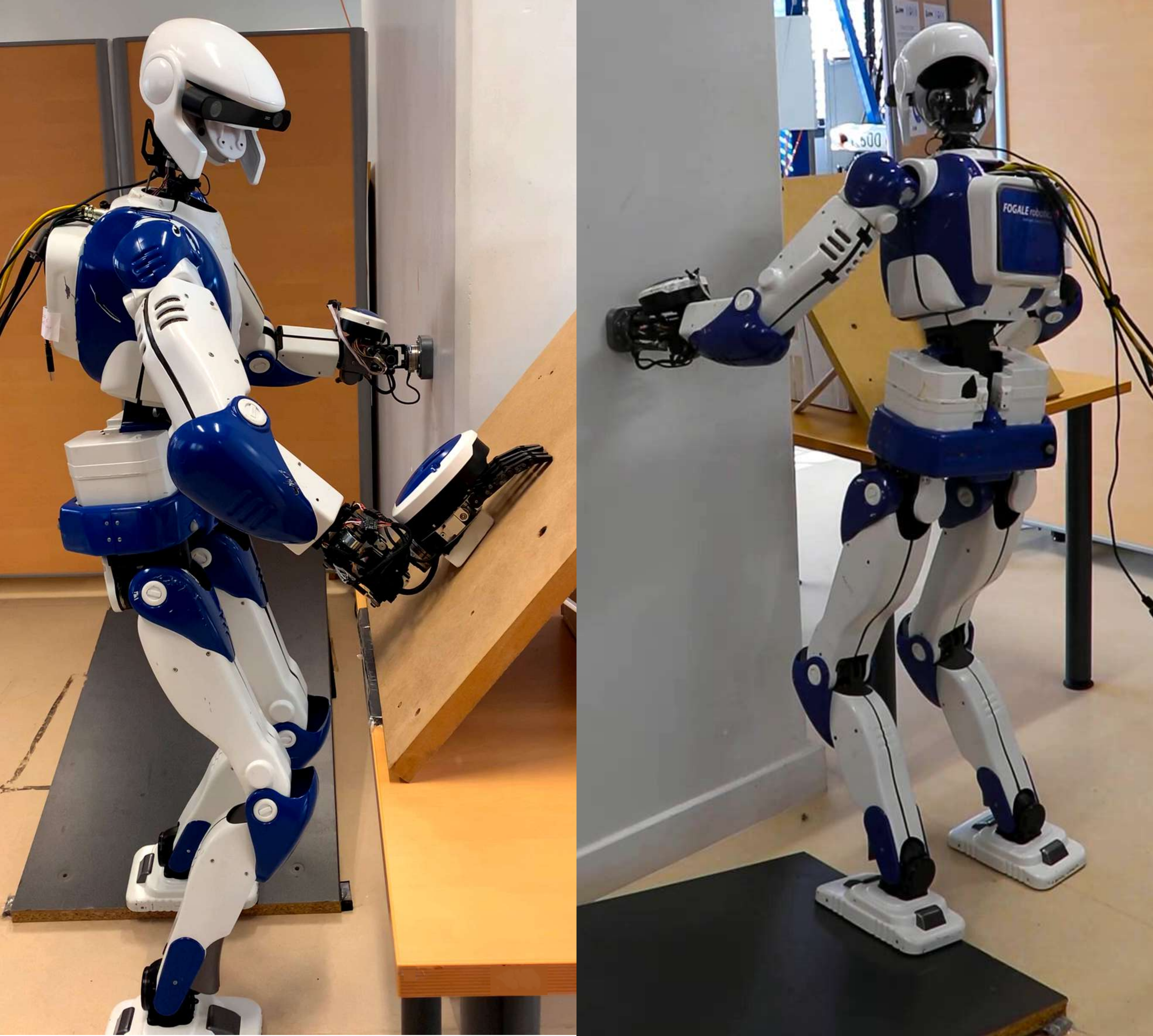}
	\caption{Non-coplanar (four) multi-sliding and fixed contacts and on-demand switches between them (including contact removal) with HRP-4.}
	\label{firstPageFig}
\end{figure}

Sliding contacts under active balance is challenging in humanoid robots. There are however successful achievements in specific tasks such as foot shuffling~\cite{kojima2015iros,or2019icra}; slip-turns and maneuvers by two feet contacts~\cite{kojima2017ral,miura2013tro,koeda2011icra}. Sliding contacts forces must be controlled to be exactly on their friction cone~\cite{trinkle1997}. We have addressed this problem in~\cite{samadi2020icra}, where we considered a mix of sliding and fixed multi-contact scenarios. The CoM support area (CSA) is derived analytically when the fixed contacts are coplanar. To overcome this limitation and extend the balance criteria to be used in control, we propose a novel formulation for the computation of the CoM position and the contact wrenches using the Chebyshev center.

Multi-sliding and fixed contacts are considered as in Fig.~\ref{firstPageFig}. We implemented an online friction estimation to adjust the real coefficient of friction during sliding. The output reference CoM position and contact wrench distribution are achieved using task-space whole-body admittance control~\cite{bouyarmane2019tro}.

\section{Background} 
\label{Sec_StateOfTheArt}

Two main ingredients shall be considered for a stable dynamic balance of humanoid robots~\cite{featherstone2008book}: (i) a good control of the robot/environment interaction forces, and (ii) a good control of the dynamic motion, governed by Newton-Euler equations, contact constraints and balancing of the external wrench. Standing or walking with a stable balance on flat terrains~\cite{wieber2006humanoids,brasseur2015humanoids} can be achieved by maintaining the ZMP inside its contacts support area~\cite{senoo2017robotica}; the latter being the convex hull of all the contact points. Whereas for non-coplanar multi-contact postures and motions, sustaining stable balance is less trivial to achieve.

\subsection{Related Works and Contribution} 
\label{Cutting-edge Investigations}

Enforcing dynamic balance of humanoid robots in multi-contact non-coplanar configurations can be addressed with one of the following inclusions:
\begin{itemize}
	\item constraining the gravito-inertial wrench within the GIWC, e.g.~\cite{caron2015rss};
	\item for a given CoM acceleration convex set and contact friction cones, build a safe region (convex polyhedron) for the CoM, e.g.~\cite{audren2018tro};
	\item for a given CoM convex set and contact friction cones, built a safe region (convex polyhedral cone) within which the CoM acceleration shall safely lie, e.g.~\cite{caron2016humanoids}. 
\end{itemize}
Efficient and fast computation of these balance equilibrium regions has been thoroughly investigated.

In~\cite{caron2015rss} multi-contact balance is fulfilled by computing the GIWC for each stance. In this approach, there is no need to update the GIWC for fixed contact stances. However, in the case of moving or sliding contacts, GIWC needs to be re-computed at each iteration. Unfortunately, since calculating the GIWC is computationally expensive, it is not possible to use this method in closed-loop control. In a recent work,~\cite{abi-faraji2019ral} challenged this shortcoming by specifying two different set of contacts in multi-contact motion settings: \emph{interacting} (hence moving) contacts, and \emph{balancing} contacts that are chosen and constrained to be fixed (i.e. static contacts). Subsequently, only balancing contacts are used to compute the GIWC. Other contacts are considered as external forces induced by tasks to be balanced altogether with the robot dynamics. This approach is appropriate when the external contacts are concerned with holding a free-floating object. However,~\cite{abi-faraji2019ral} is limiting in all the other cases where task-induced contacts (including the moving ones) can be exploited for balance (e.g. pushing an object, sliding...). This is because with~\cite{abi-faraji2019ral}, task contacts won't be allowed to contribute to balance and this is clearly not the way, we humans perform. By categorizing contacts this way (even if switches are possible), the multi-contact GIWC pre-calculation is made on a subset of contact (those fixed) for balance, hence relatively tractable; yet it excludes all other contacts that could eventually contribute to balance.

Computation efficiency also exists for the CoM-support regions~\cite{audren2018tro}. It was leveraged using 3D morphing techniques between two regions~\cite{audren2016humanoids} but without guarantee on the balance validity all along the morphing shape pathway between the two regions. To control a humanoid in multi-contact,~\cite{sentis2010tro} used the virtual-linkage model. Their method consists in constructing a multi-contact CoM area as the envelope of valid points.
In our recent previous work~\cite{samadi2020icra}, we introduced an analytical solution to compute the CoM-support area in real-time. However, these developments apply only when the fixed contacts are coplanar.

We propose an alternative formulation that allows balance criteria to be used in closed-loop control, that applies to moving/sliding and fixed contacts. It also permits on-the-fly switching between these contact modes. Our approach do not separate contacts for balance from those of the tasks, when the latter can contribute to balance (and \emph{vice versa}, i.e. balance contribute to achieving the task). Our approach distinguishes from existing work as follows:
\begin{itemize}
	\item No need for constructing or pre-computing explicitly the multi-contact balance region (GIWC);
	\item Fast computation performances enabling real-time closed-loop control;
	\item Calculating CoM position and wrench distribution of contacts using the Chebyshev center;
	\item Covering all types of contact modes (e.g. multi-sliding contacts) without any of the limitations pointed in~\cite{samadi2020icra};
	\item contrary to~\cite{abi-faraji2019ral}, we do not exclude moving/tasks contacts to contribute to dynamic balance stability.
\end{itemize}

\subsection{Centroidal Model} \label{Basics}

We consider motion scenarios without locomotion. Static equilibrium of the robot is represented by Newton-Euler equation for $l$ limbs in contact with the environment:
\begin{equation}
\bfw^g + \sum_{i=1}^l {\bfw_{i}^c} = 0 \label{NE}
\end{equation}
where $\bfw^g\in \mathbb{R}^6$ and $\bfw_{i}^c \in \mathbb{R}^6$ are gravity wrench and the $i^{\text{th}}$ contact wrench in the world frame respectively, and $ \bfw = [\bff \ \ \bftau ]^T$. Gravity and contact wrenches are specified in the following form:
\begin{equation}
\bfw^g=\begin{bmatrix}
\bff^g \\ \bfc \times \bff^g
\end{bmatrix} \;\;\;
\bfw_{i}^c = \begin{bmatrix}
\bff_i^c \\ \bfp_i \times \bff_i^c + \bftau_i^c
\end{bmatrix}
\label{GravityContactWrenches}
\end{equation}
where $\bff^g = [0 \ \ 0 \ \ mg]^T$, $\bfc = [\bfc_x \ \ \bfc_y \ \ \bfc_z]^T$ the position of the CoM, and contact points $\bfp_i$ are given with respect to a global frame. Note that the contact wrenches are mapped from the local to the global frame by rotation matrix $[\bfR]_i \in \bbR^{6\times6}$:
\begin{equation}
\bfw_i^c = [\bfR]_i \ ^l\bfw_i^c 
\end{equation}
where $^l\circ$ denotes the local frame of the contact point.

We consider the following state variables to be computed at each iteration:
\begin{align}
\bfY = [\bfc \ \ \mathbf{\calW}_1 \ \ \mathbf{\calW}_2 \ \ \hdots \ \  \mathbf{\calW}_l ] \label{DecisionVariables_Y}
\end{align}
where $ \mathbf{\calW}_i$ is the $i$-th contact wrench $\bfw_i^c$.

\subsubsection{Sliding and equality bounds on contact forces} 
\label{EqualityConstraintsSubsection}

For every contact of the robot, we determine the wrench. Equations~\eqref{GravityContactWrenches} can be re-written as:
\begin{align}
\bfw^g &= \bfA^g \bfc - \bfb^g \label{EqualityMass}\\
\bfw_{i}^c &= \bfA_{i}^c \mathbf{\calW}_i \;\;\; \text{ where } i=1,\ldots,l \label{EqualityContacts}
\end{align}
On the other hand, the desired sliding comes with additional equality constraints as stated in~\cite{samadi2020icra}. By convention, the normal contact force ($f_z^c$) for sliding is aligned to the normal of the local surface frame. The dynamic friction coefficient ($\bfmu$) of each contact, is estimated online during sliding. The force along other axes ($f_x^c$ and $f_y^c$) of the tangent space is derived from the pre-defined velocity and trajectory of the sliding motion. Let,
\begin{equation}
^l\bff_i = \begin{bmatrix}
f_{x,i} & f_{y,i} & f_{z,i}
\end{bmatrix}^T \;\;\; i = 1,\ldots,s,
\end{equation} be the sliding contact forces in the local frame, $s$ being the number of sliding contacts. We can write in a matrix form
\begin{equation}
^l\bff_k = \begin{bmatrix}
0&0&\mu_{x,k} \\
0&0&\mu_{y,k} \\
0&0&0
\end{bmatrix} \ ^l\bff_k + 
\begin{bmatrix}
0 \\
0 \\
f_{z,k}
\end{bmatrix}\label{frictionCoef}
\end{equation} or,
\begin{equation}
\begin{bmatrix}
1&0&-\mu_{x,k} \\
0&1&-\mu_{y,k} \\
0&0&1
\end{bmatrix} \ ^l\bff_k = 
\begin{bmatrix}
0 \\
0 \\
f_{z,k}
\end{bmatrix}
\end{equation} giving the $k^{\text{th}}$ sliding contact velocity $\bfv_k = v_{x,k}i + v_{y,k}j$, let $\alpha_{x,k} = \frac{v_{x,k}}{\norm{\bfv_k}}$ and $\alpha_{y,k} = \frac{v_{y,k}}{\norm{\bfv_k}}$, from which $\mu_{x,k} = \mu_k\alpha_{x,k}$ and $\mu_{y,k} = \mu_k\alpha_{y,k}$; $\mu_{k}$ is the $k^{\text{th}}$ dynamic friction coefficient.

The above equation writes in a compact form:
\begin{equation}
\bfC \ ^l\bff_k = \calK
\end{equation}
$\text{det}(\bfC)=1$, $\bfC$ is an invertible matrix, hence:
\begin{equation}
^l\bff_k = \bfC^{-1}\calK
\end{equation}
Also, by using matrix transforms, we change the coordinate to the global frame:
\begin{equation}
\bff_k^c = [\bfR_k]_{3\times3} \bfC^{-1}\calK
\end{equation}
which can be written using the selection matrix $[S]_{3\times6} = [I_{3\times3} \ \ 0_{3\times3}]$ as:
\begin{equation}
[S]\bfw_k^c = [\bfR_k]_{3\times3} \bfC^{-1}\calK \label{SlidingEqualityConstraint}
\end{equation} 

Equation~\eqref{SlidingEqualityConstraint} is embedded to the controller along with eqs.~\eqref{EqualityMass}~and~\eqref{EqualityContacts}, and can be written as:
\begin{align}
\bfA_i^{sl} \calW_i - \bfb^{sl}_i = 0 \text{; } i = 1, \ldots , s
\end{align}
This equation is applied to the $s$ sliding contacts w.r.t their desired sliding forces. Hence, in this equation $\mathbf{\calW}_i$ refers to sliding contacts only.
\subsubsection{Fixed contacts} 
\label{InequalityConstraintsSubsection}
The non-sliding condition of contacts is fulfilled when the contact force lies strictly within the friction cone. Linearized equations for non-sliding conditions were thoroughly discussed in~\cite{caron2015thesis}. The following equations are sufficient conditions to avoid slippage and tilting of the contacts in all $\mathbb{R}^6$ coordinate of the local frame:
\begin{equation}
\begin{split}
\mid f_x  \mid \leqslant \mu f_z \;,\; \mid f_y \mid \leqslant \mu f_z  \;,\;  f_z^{\min} \leqslant f_z  \leqslant f_z^{\min} \\ 
\mid \tau_x \mid  \leqslant  \sigma_y f_z \;,\;  \mid \tau_y \mid \leqslant  \sigma_x f_z  \;,\; \tau_z^{\min} \leqslant \tau_z  \leqslant \tau_z^{\min} 
\end{split}
\label{FrictionCone}
\end{equation} where $\sigma_*$ are scalars defined in~\cite{caron2015thesis}. 

By considering $n$ non-sliding contacts, the inequality constraints are in the following form:
\begin{subequations}
	\begin{align}
	\bfUpsilon_{i}^{ub} \calW_i &\leqslant \textbf{h}_{i}^{ub} \;\;\; i=1,\ldots,n \\
	\bfUpsilon_{i}^{lb} \calW_i &\leqslant \textbf{h}_{i}^{lb} \;\;\;  i=1,\ldots,n
	\end{align}
\end{subequations}
where indexes $ub$ and $lb$ shows the upper and lower bounds. $\bfUpsilon_i$ matrices and $\bfh_i$ vectors are introduced in~\cite{samadi2020icra} for both sliding and fixed contacts. Also, $\bfPsi_i$ enforces inequality constraints on $\tau_z$ element of wrenches:
\begin{subequations}
	\begin{align}
	\bfPsi_{i}^{ub} \calW_i &\leqslant 0_{4\times 1} \;\;\; i=1,\ldots,n \\
	\bfPsi_{i}^{lb} \calW_i &\leqslant 0_{4\times 1} \;\;\; i=1,\ldots,n
	\end{align}
\end{subequations}

For the sliding contacts, the same inequality constraints are implemented only on the torque of the wrench to avoid the tilting of the sliding contacts.

\section{Optimal Control Framework} \label{Sec_Planner}

In this section, we propose the implementation of the Chebyshev center theorem~\cite{beck2007siam} in a two-level control framework. Inspired from~\cite{collette2007humanoids,samadi2020icra} such an implementation can be formulated as a first-level quadratic program (QP) that outcomes in real-time, the CoM position and contact wrenches distribution (together with the Chebyshev center and its radius). These are then integrated as tasks objectives or constraints in a running whole-body task-space controller formulated as a second-level QP~\cite{bouyarmane2019tro} (see later Section~\ref{Sec_WBQP}).
\begin{figure}[!htb]
	\centering
	\includegraphics[width=0.7\columnwidth]{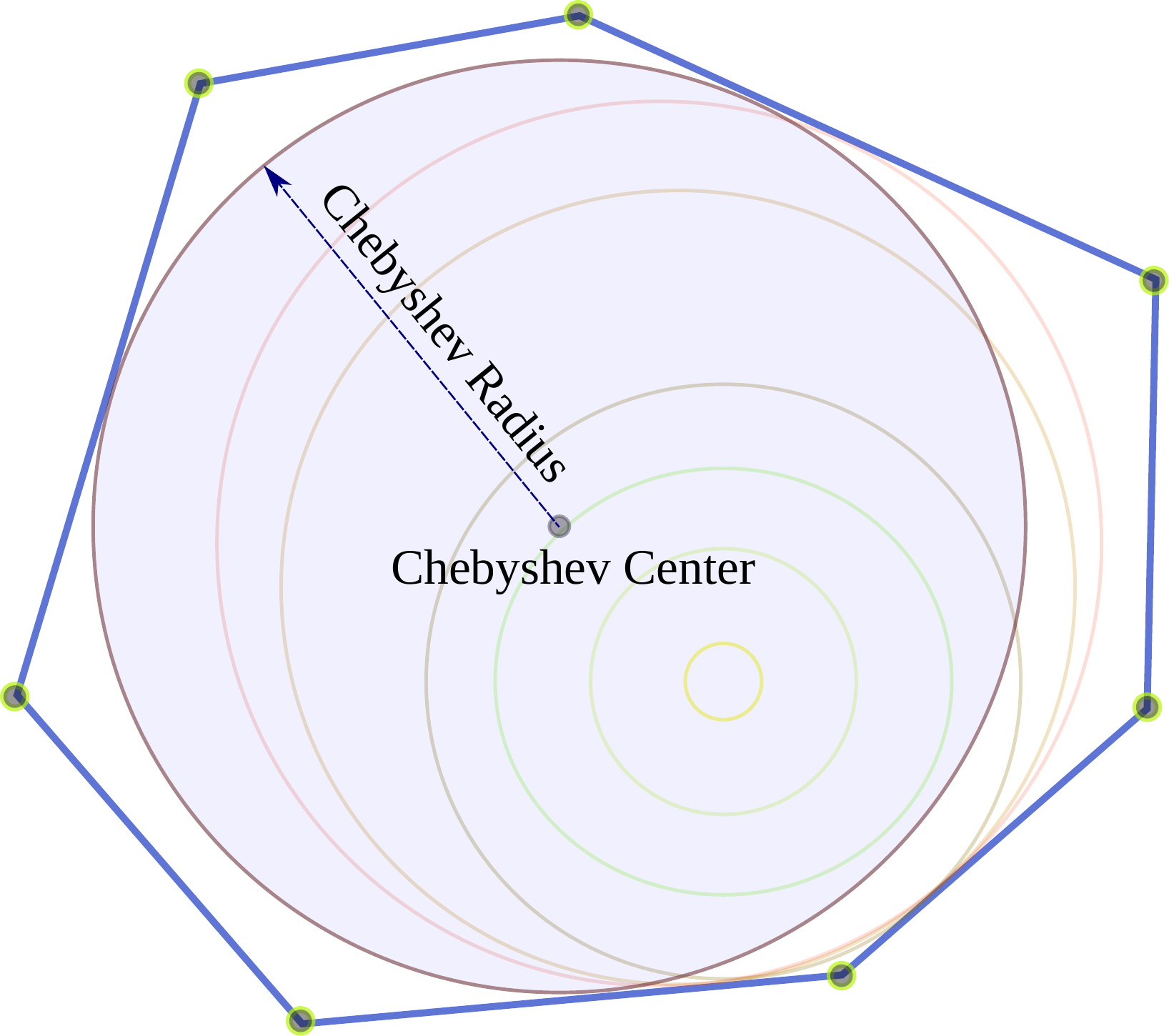}
	\caption{Chebyshev circle enclosed in a polygon.}
	\label{ChebyshevCircle}
\end{figure}

\subsection{Chebyshev Center} 
\label{ChebyshevCenterSubsection}

Let $U$ be a set of inequality constraints describing a bounded polygon. The Chebyshev center of $U$ is the \emph{center of the largest enclosure circle}~\cite{amir1985}. Figure~\ref{ChebyshevCircle} illustrates the Chebyshev center and radius of a given polygon. To compute the Chebyshev center, we do not need to build the vertices of the polygon (i.e. no need for the v-representation of $U$).

For a point, $x$ in the interior of $U$, the Chebyshev center $\hat{x}$ is a point among all possible $x$ such that~\cite{knitter1988}:
\begin{equation}
\arg \min_x \max_{x \in U} || x - \hat{x}|| ^2 
\end{equation}

This applies to any bounded non-empty \emph{convex set}
\begin{equation}
\bfU = \{ \bfx \ | \ f_i(\bfx) \leqslant 0 ; i = 1,\ldots,n\}, \label{cheb_U}
\end{equation}
where $\bfx$ is a vector of a given dimension, and $n$ the number of inequalities. These inequalities are valid for all points inside the convex set. Hence, constituent points of the Chebyshev ball enclosed in the convex set, with center $\hat{x}$ and radius $r$ fulfill the inequalities too. The idea is to compute the largest Chebyshev ball which satisfies the convex condition~\eqref{cheb_U}:
\begin{subequations}
\begin{align}
	\max \ & r \\
	\text{w.r.t} \ & \sup_{\norm{\bfa} \leqslant 1} f_{i= 1,\ldots,n}(\bfx + r \bfa) \leqslant 0 \label{Cheb_convex}
\end{align}
\end{subequations}
and the supremum form of~\eqref{Cheb_convex} allows accessing all points within the ball according to the variable $\bfa$.
\begin{figure}[!htb]
	\centering
	\includegraphics[width=0.95\columnwidth]{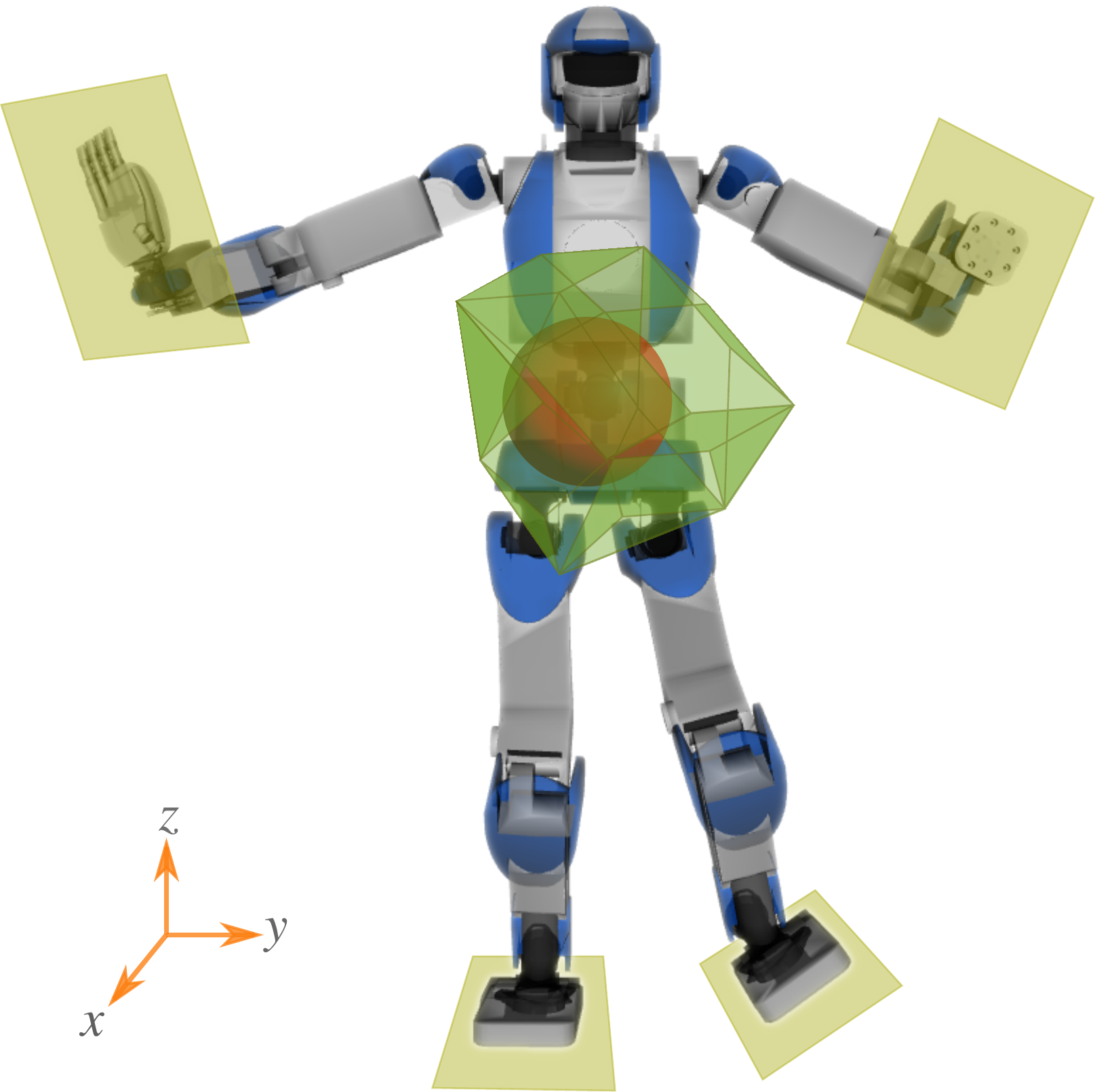}
	\caption{Illustration of the Chebyshev center of an enclosed Euclidean ball within a multi-contact balance GIWC.} 
	\label{Chebyshev}
\end{figure}

For a polytope, the inequalities are in the form of $\{\bfalpha_i^T \bfx - \bfbeta_i \leqslant \mathbf{0}, \; i = 1,\ldots,n\}$. So the condition~\eqref{Cheb_convex} is modified and the optimization problem is\footnotemark:
\footnotetext{By replacing the inequality of the \emph{polytope} into~\eqref{Cheb_convex}, the term $\bfalpha_i^T r \bfa$ appears. Supremum of this term will provide $r \norm{\bfalpha_i}_*$. Note that the trace norm, $\norm{}_*$, is introduced for computing the norm of the matrices. In case of vectors, one can use the simple $l_{1}$ or $l_{2}$ vector norms.} 
\begin{subequations}
	\begin{align}
	\max \ & r \\
	\text{w.r.t} \ &  \bfalpha_i^T \bfx + r \norm{\bfalpha_i} - \bfbeta_i \leqslant \mathbf{0}\;\;\; i = 1,\ldots,n \label{Cheb_polyhedronIneq} \\ 
	& r \geqslant \mathbf{0}
	\end{align} \label{Cheb_polyhedron}
\end{subequations}

Figure~\ref{Chebyshev} illustrates the link between the above generalities concerning the Chebyshev center and humanoid balance. The polyhedron illustrates the GIWC, that can be obtained as in e.g.~\cite{abi-faraji2019ral}. The benefit of computing the Chebyshev center, instead of the GIWC, is undeniably the speed. The price to pay for the speed is the conservativeness of the balance region as can clearly be seen in Fig.~\ref{Chebyshev}.  

\subsection{Chebyshev Quadratic Programming} \label{QP}

Recall from~\eqref{DecisionVariables_Y} that the decision variables are noted as $\bfY = [\bfc\ \ \calW_{i = 1,\ldots,l}]^T$ where $\bfY \in \bbR^{3+6k}$. The QP formulation is structured based on the equality and inequality constraints introduced in section~\ref{Basics} expressed as:
\begin{subequations}
	\begin{align}
	\bfA \bfY &= \bfb \label{eq} \\
	\bfG \bfY &\leqslant \bfh \label{ineq}
	\end{align}\label{QP_firstVersion}
\end{subequations} whereas:
\begin{align}
\bfA = \begin{bmatrix}
\bfA^g                  && \bfA^c \\
\mathbf{0}_{6n \times 3} && \bfA^{sl}
\end{bmatrix}_{(6+6n)\times(3+6k)} \label{A_matrix}
\end{align}
and the matrices $\bfA^c$ and $\bfA^{sl}$ are defined as:
\begin{align}
\bfA^c &= \begin{bmatrix}
\bfA^c_1 && \dots && \bfA^c_k
\end{bmatrix}_{6\times6k} \\
\bfA^{sl} &= [\diag(\bfA_{i}^{sl})] \bfrho^{sl}; \ i = 1,\ldots, n
\end{align}

The equation above contains a selection matrix $\bfrho^{sl}_{6n \times 6k}$ for the elements related to the sliding contacts in the $\bfA^{sl}$ matrix and $\diag()$ refers to the diagonal matrix structure.
The rest of the parameters of~\eqref{QP_firstVersion} are defined as:
\begin{align}
\bfb &= \begin{bmatrix}
-\bfb^g && \bfb^{sl}_1 && \dots && \bfb^{sl}_n
\end{bmatrix}^T_{1 \times 7n} \\
\bfG &= [\diag(\bfUpsilon_i,\bfPsi_i)]_{20k \times 10k} \bfrho^{G} ; \ i = 1,\ldots,k \\
\bfh &= \begin{bmatrix}
\bfh^{ub}_i && \bfh^{lb}_i && \mathbf{0}_{1 \times 8k}
\end{bmatrix}^T_{1 \times 20k} ; \ i = 1,\ldots,k
\end{align}
and $\bfrho^G_{20k \times 3+6k}$ is the selection matrix for the $\bfG$ matrix.

For maximizing the radius and selecting the optimal Chebyshev center, inequality constraint~\eqref{ineq} is modified in the following form, based on~\eqref{Cheb_polyhedronIneq}, in presence of Chebyshev radius $r$:
\begin{equation}
 \bfG \bfY+r \bfxi \leqslant \bfh \label{Cheb_ineq}
\end{equation}
where $\bfxi \in \bbR^{20k}$ is a vector which consists of the norm of the rows of $\bfG$ matrix separately:
\begin{equation}
\bfxi_j = \norm{\bfG(j,:)} \;\;\;\;\; j=1,\ldots,20k
\end{equation}
and the operator $(j,:)$ shows the $j^{\text{th}}$ row of the matrix. The vector $\bfxi$ in~\eqref{Cheb_ineq} is equivalent to the vector $\norm{\bfalpha}$ where the $i$-th element of the vector $\bfalpha (i) = \bfalpha_i \; \forall i \in \{1,\ldots,n\}$ in~\eqref{Cheb_polyhedronIneq}. This way, we aim at maximizing the Chebyshev radius by taking into account all inequalities without the need to compute the GIWC. 
The decision variables of this first-level QP are the position of CoM, wrench distribution, and $r$. To include the Chebyshev radius within the decision variables, matrices are modified as follows:
\noindent
\begin{multicols}{2}
	\noindent
	\begin{align}
	\bfX &= [\bfY \ \ r]^T \\
	\bfh^* &= [\bfh \ \ 0]^T
	\end{align}
	\begin{align}
	\bfA^* &= [\bfA \ \ 0] \\
	\bfG^* &=\begin{bmatrix}
	\bfG & \bfxi \\
	\mathbf{0} & -1
	\end{bmatrix}
	\end{align}
\end{multicols}
\noindent
Maximizing the Chebyshev radius is equivalent to minimizing $(-r)$. Other objectives of the QP can be:
\begin{itemize}
	\item Setting the current position of the CoM as a target for the next iteration (smoothing CoM trajectory)
	\item Minimizing the wrench distribution: sharing the load on non-constrained (contact) force.
\end{itemize}
These objectives can be defined through $\bfX_{\text{des}} = [\bfY_{\text{des}} \ \ r_{\text{des}} ]$, and the optimization framework writes:
\begin{subequations}
	\begin{align}
	\min_{\bfX} \ \ \norm{\bfX-\bfX_{\text{des}}}^2-r \ &\equiv \ \frac{1}{2} \bfX^T \bfP \bfX + \bfq^T \bfX\\
	\bfG^* \bfX &\leqslant \bfh^* \\
	\bfA^* \bfX &= \bfb
	\end{align} \label{minimization}
\end{subequations}
where
\begin{multicols}{2}
	\noindent
	\begin{align}
	\bfq &= [\bfY_{\text{des}} \ \ -1]
	\end{align}
	\begin{align}
	\bfP &= 2\calQ
	\end{align}
\end{multicols}
\noindent
We prioritize the solution of the QP by using a weight matrix $\calQ_{(4+6k) \times (4+6k)}$ with chosen weights on the corresponding decision variables (the diagonal elements). However, we do not set any target for $r$. Therefore, we need to adopt a low computation weight for $r_{\text{des}}$.
This QP problem, named Chebyshev QP for the rest of the paper, provides the optimal position of CoM and wrench distribution by maximizing the Chebyshev radius.

\subsection{Calculation of the Range of Contact Wrench and CoM Position}

Chebyshev radius is the output of the minimization problem~\eqref{minimization}. However, this radius indicates a range that applies to all contact wrenches. We can additionally calculate the range of the gravito-inertial wrench within the GIWC based on the computed radius. Note that the contact wrench cone (CWC) is simply the opposite of the GIWC. So, the computation for range accounts for both whole-body contact and gravito-inertial wrenches.
Assuming a set of contact wrenches, expressed at CoM, within their respective contact wrench cones $\bfw_i \in \bfC_i$ for $ i=1,\ldots,l$, and representation of Chebyshev method~\eqref{Cheb_convex}, we have
\begin{align}
	\bfw_i + r \bfa \in \bfC_i
\end{align}
Regarding the sum of contact wrenches and their ranges ($r \bfa$) we can write
\begin{align}
	\sum_{i =1}^{l}\bfw_i + l \times r \bfa \in \bfC_1 \oplus \ldots \oplus \bfC_l
\end{align}
where $\oplus$ denotes the Minkowski sum of the contact wrenches cones which is the explicit definition of the CWC. So, the range for the valid whole body contact and gravito-inertial wrenches is the sphere with the radius of $r_{w} = l \times r$.
\subsection{Online Estimation of the Friction Coefficient}

Implementing the sliding motion of contact needs a good estimation of the friction coefficient. The latter is considered in~\eqref{frictionCoef} and has a major role in matrix $\bfA^{sl}$ of~\eqref{A_matrix}. Dynamic and static friction coefficients are defined as a property of a pair of surfaces in contact and can not be known intrinsically. Let $^lf_i^{c,\text{tan}}$ and $^lf_i^{c,\text{nor}}$ be the tangential and normal forces respectively in the local frame of the $i$-th contact surface, the norm of the Coulomb friction equation~\cite{williamson1996} leads us to the following formulation:
\begin{align}
\norm{^lf_i^{c,\text{tan}}}_2 = \mu^{\text{mes}} \mid ^lf_i^{c,\text{nor}} \mid 
\end{align}
where $^lf_i^{c,\text{tan}} = [^lf_{x,i}^{c} \ ^lf_{y,i}^{c}]^T$ and $\mu^{\text{mes}}$ is the calculated friction coefficient from the measured local forces of the force sensors:
\begin{align}
\mu^{\text{mes}} = \frac{\sqrt{{^lf_{x,i}^{c}}^2 + {^lf_{y,i}^{c}}^2}}{\mid ^lf_{z,i}^{c} \mid}
\end{align}

To minimize the effect of the force measurements noise, we also apply a simple filter with $0 \leqslant \gamma \leqslant 1$ and calculate the filtered friction coefficient at each time iteration $t$:
\begin{align}
\mu^{\text{filt}}_t = \gamma \mu^{\text{mes}}_{t-1} + (1 - \gamma) \mu^{\text{mes}}_t  \label{filter}
\end{align}

\section{Whole-body Admittance Controller} \label{Sec_WBQP}

\begin{figure}[h!]
	\centering
	\includegraphics[width=\columnwidth]{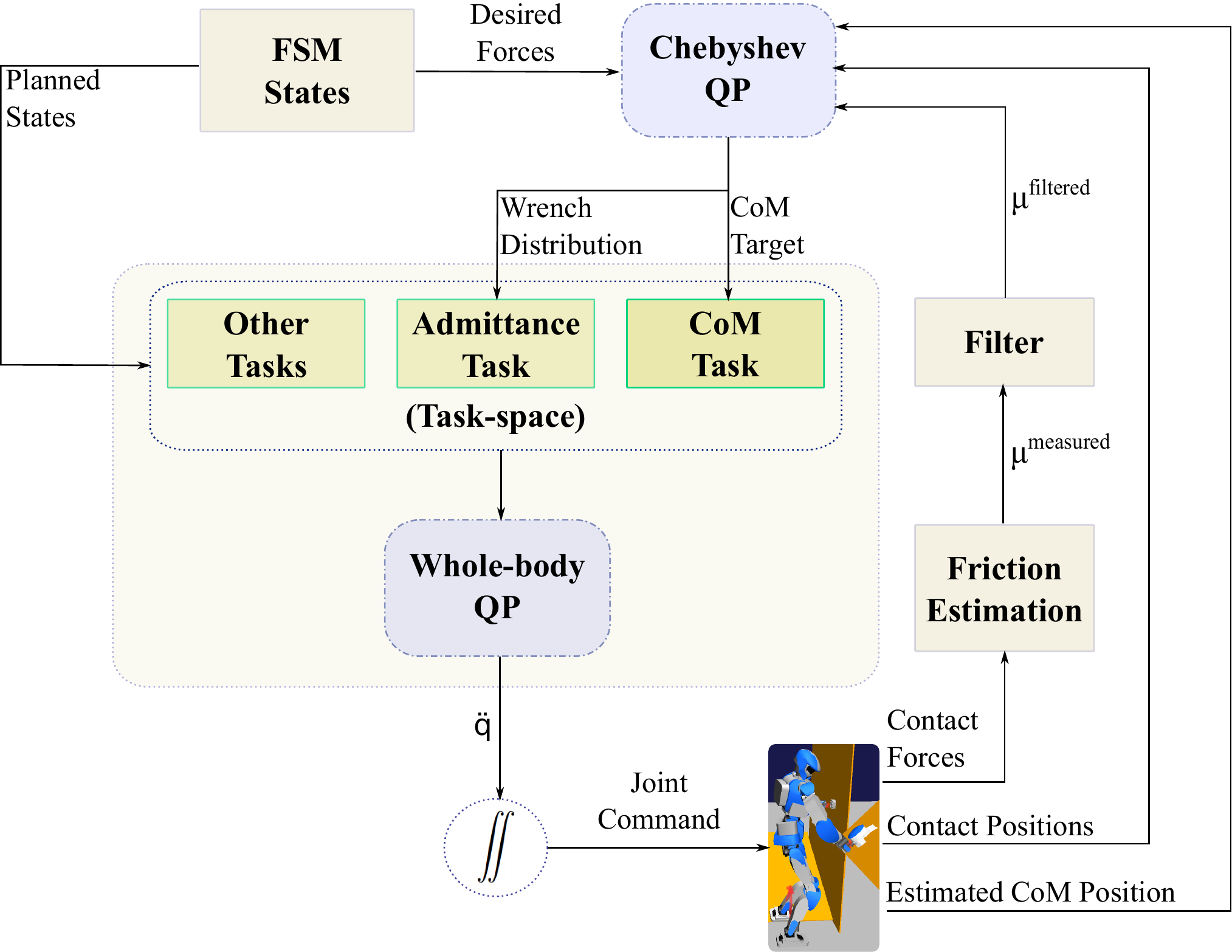} 
	\caption{Schematic of the overall task-space control.}
	\label{schematic}
\end{figure}

In this section, we present the overall task-space whole-body admittance controller as shown in Fig.~\ref{schematic}. For each scenario, a Finite State Machine (FSM) is designed/planned. Each state of the FSM, defines a set of tasks and constraints for the controller as well as desired positions, forces, and sliding conditions for each contact, if any.

The Chebyshev QP computes optimal contact wrenches and CoM position depending on the current state of the robot and FSM targets. These optimal wrenches and CoM position are then used as objectives for tasks driven by our whole-body QP (WBQP) framework. Each task is formulated as a cost function and/or eventually associated constraints that depend on joint acceleration (or torque for torque-based humanoids) and contact forces. The resulting joint acceleration is integrated twice (or torques) and sent as a target for the robot actuators' low-level controllers, see~\cite{bouyarmane2019tro}.

The main tasks we used in our experiments are the following: (i) CoM task, to eventually track the optimal CoM position, (ii) Admittance tasks for the desired wrench which map wrench error to contact surface velocity which in turn is used as a target for an end effector trajectory task.

\begin{figure}[!t] 
	\subfigure[]{
		\includegraphics[width=\columnwidth]{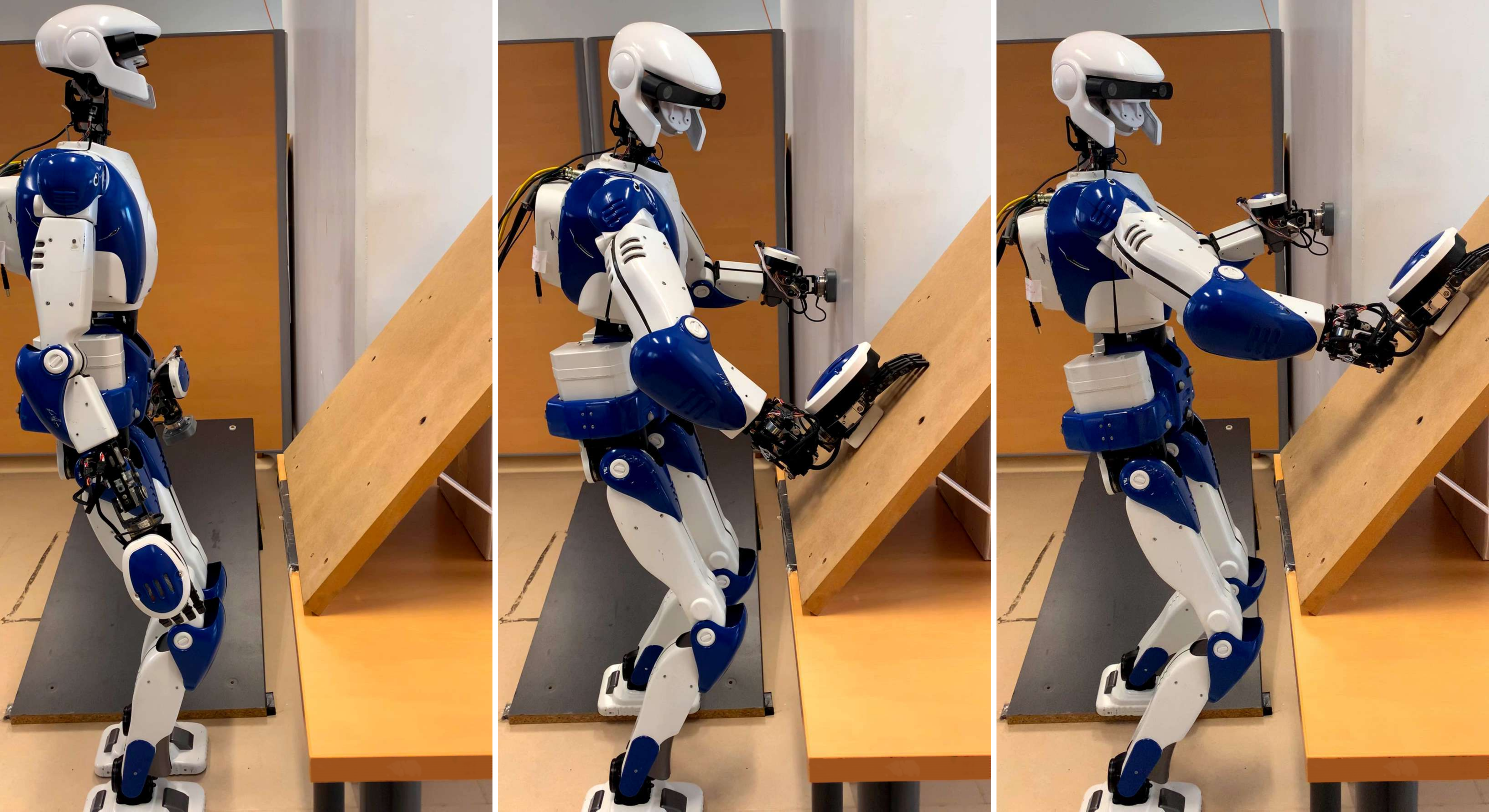}
		\label{coWipingExp}
	}
	\subfigure[]{
		\includegraphics[width=\columnwidth]{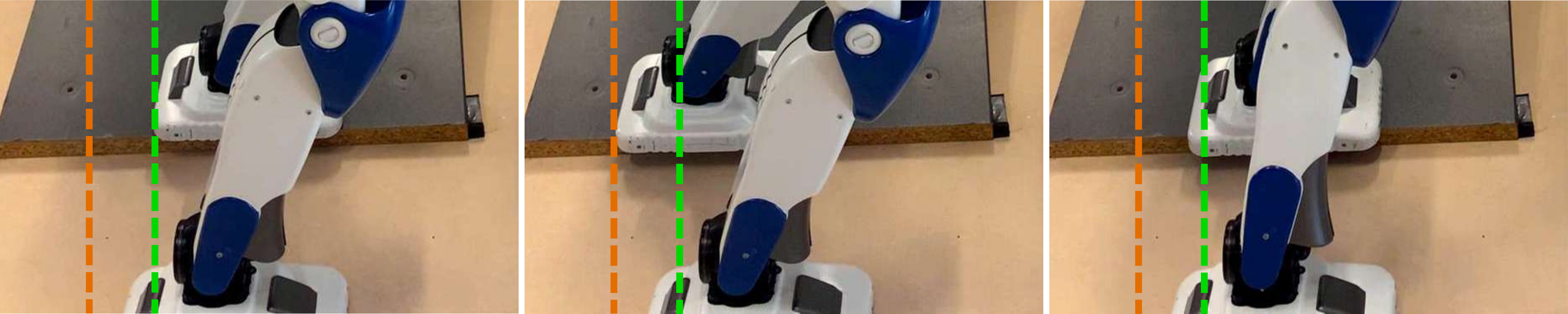}
		\label{shufflingExp}
	} 
	\caption{Stepping up the slope and (a) simultaneous sliding of both hands on non-coplanar wooden board and wall surfaces, and (b) shuffling of the left foot on a slope tilted by $20^ \circ$. w.r.t the ground in a multi-contact setting. }
	\label{experiments}
\end{figure}
\begin{figure}[t!]
	\centering
	\includegraphics[width=\columnwidth]{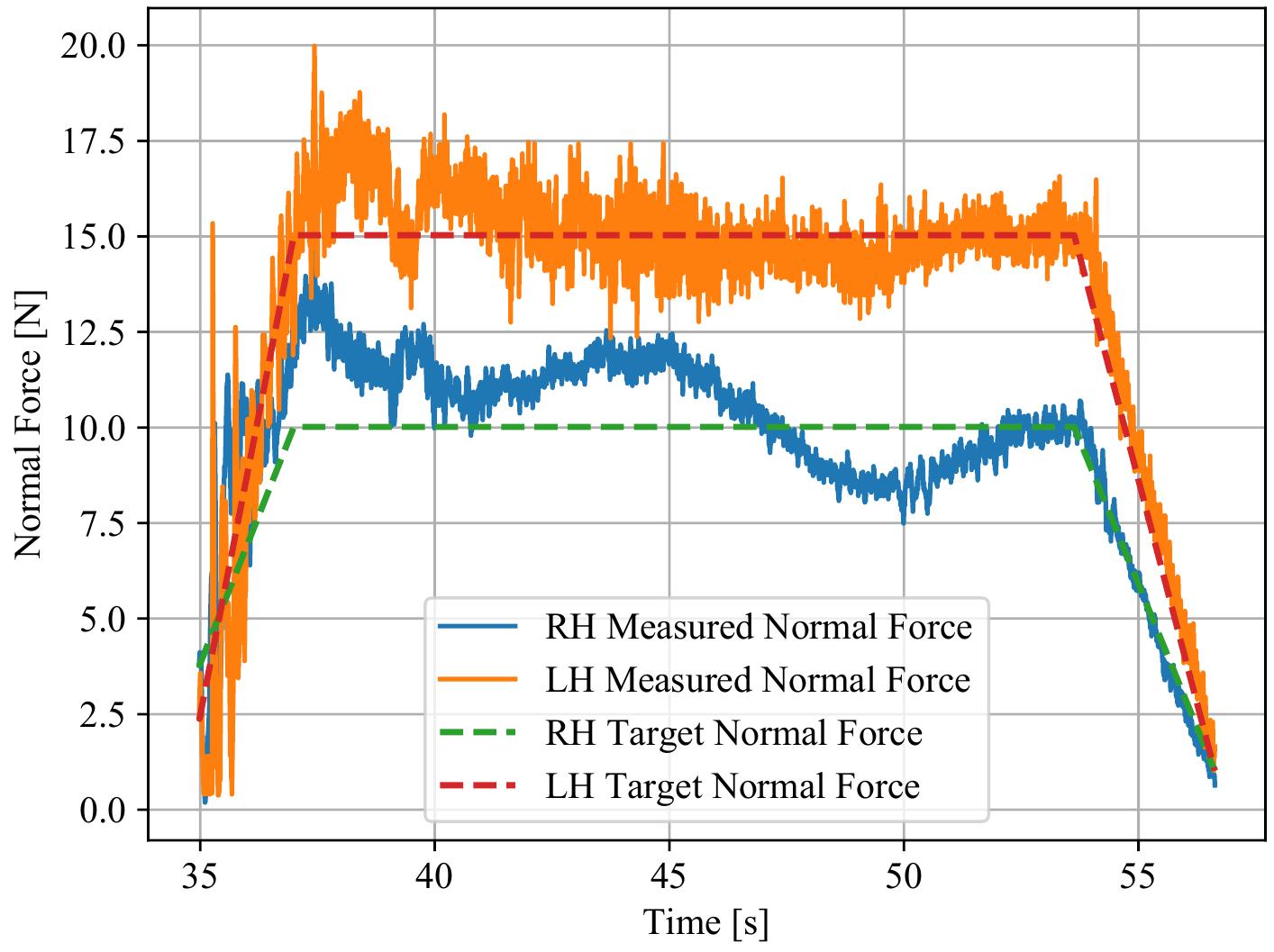} 
	\caption{Force tracking of the end-effectors during co-wiping tasks.}
	\label{cowipingForceTracking}
\end{figure} 
\section{Experimental Results} 
\label{Sec_ExperimentalResults}
In order to assess our approach, we performed experiments with the HRP-4 humanoid robot. We investigate the capabilities of our proposed optimal control framework through a multi-contact active balance scenario that exhibits:
\begin{itemize}
	\item combination and switching of fixed and sliding modes;
	\item multi-sliding contacts on non-coplanar surfaces;
	\item shuffling of the foot on a tilted surface.
\end{itemize}

\begin{figure}[t!]
	\centering
	\includegraphics[width=\columnwidth]{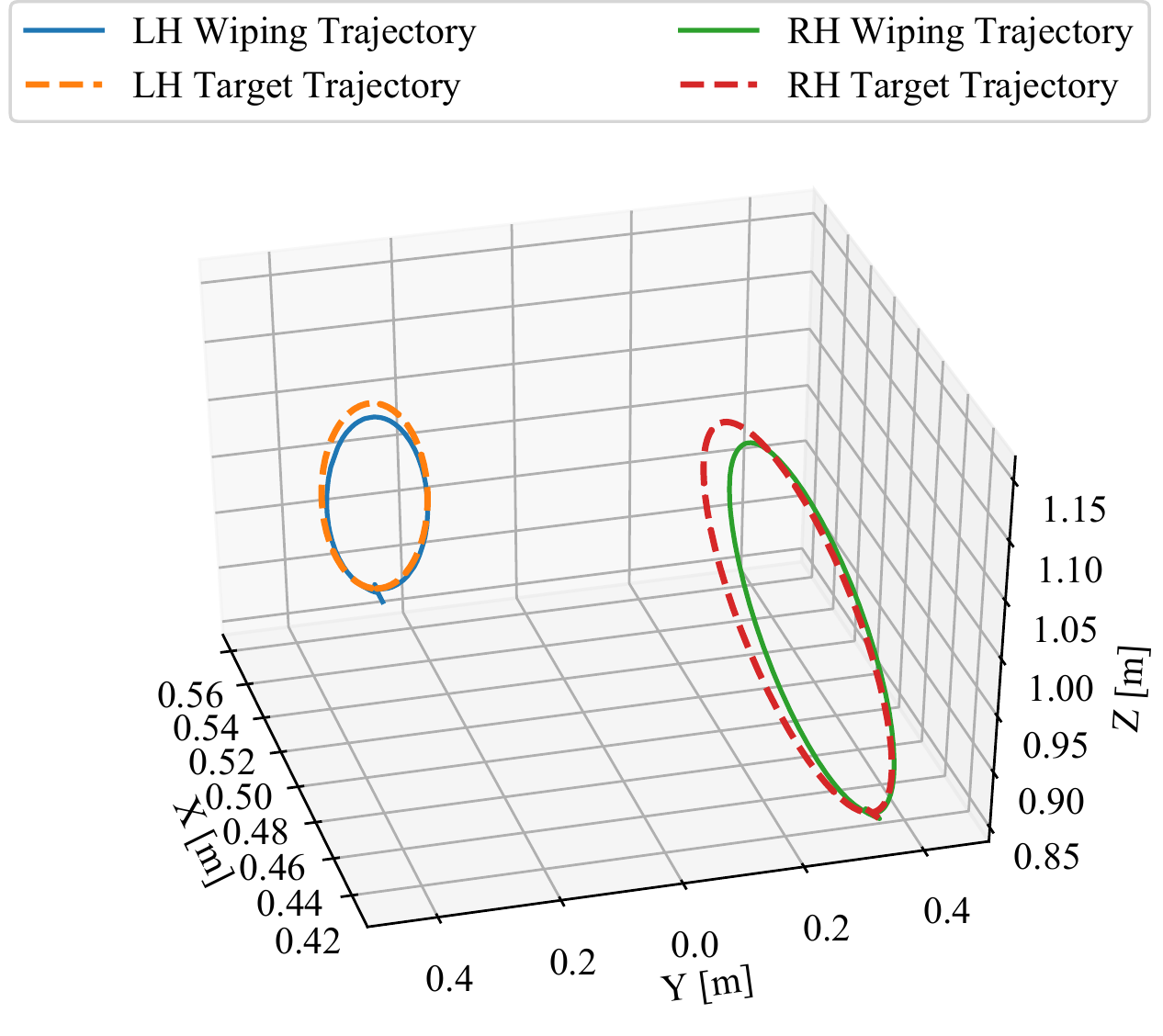}
	\caption{Target trajectory tracking of the sliding hands. The light discrepancies are due to uncertainties of  planed board and wall position w.r.t HRP-4.}
	\label{coWipeTrajectory}
\end{figure}

We prepared a non-coplanar multi-contact set-up consisting of a tilted fake wooden slope for the left foot, a tilted wooden board for the right hand, and a wall for the left hand. The right foot is on the experimental room ground. The three materials have different friction coefficients. The scenario starts by setting the HRP-4 in a half-sitting pose, both feet on the grounds. The motion begins with stepping the left foot up the slope and establishing a planar contact with the wall by the left hand. Then contact is established by the right hand on the tilted board (slope of $\simeq 50 ^ \circ$). Each contact is established while sustaining the existing ones and moving the whole body while keeping the CoM close to the suggested one by the Chebychev QP. Alike in~\cite{kheddar2019ram} and since we use guarded motion and a calibrated environment, embedded robot vision was not used this time. At this stage, we have four contacts. Then HRP-4 gets prepared for the co-wiping motion under user-specified surface normal forces: $10$~N for the right hand and $15$~N for the left hand. During co-wiping, the end effectors are tracking the desired normal force and planned trajectories of circles with a radius of $10$ and $8$~cm for right and left hands, respectively. Notice that in a multi-contact non-coplanar setting, the range of motion of HRP-4 is limited due to intrinsic kinematics closed-chain constraints. Figure~\ref{coWipingExp}~and~\ref{cowipingForceTracking} illustrate the co-wiping experiment and force tracking of the end-effectors contacts using admittance force control discussed in~\cite{bouyarmane2019tro}. The alignment of the end effectors with surfaces while establishing the contacts is guaranteed by zeroing the torques reference in $x$ and $y$ directions w.r.t the contact local frame. The trajectory tracking of end-effectors is shown in Fig.~\ref{coWipeTrajectory}.

\begin{figure}[t!]
	\centering
	\includegraphics[width=\columnwidth]{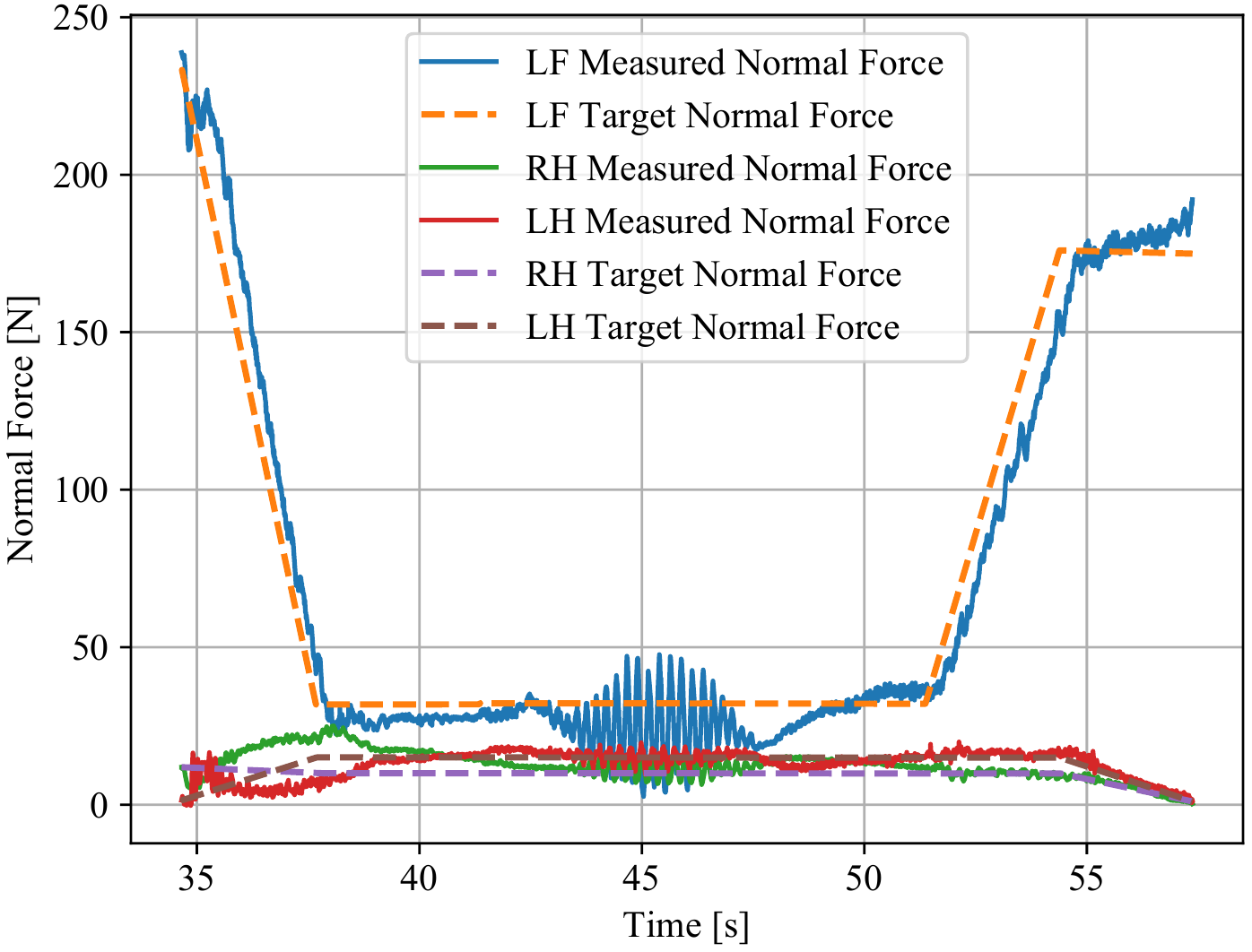}
	\caption{Normal Force tracking of hands and left foot contact during the shuffling motion. The oscillations are due to the change into the reverse direction in the shuffling motion.}
	\label{FootShuffling}
\end{figure}

Once the co-wiping tasks are achieved, both contact switch from sliding to fixed ones, and the left foot contact switches from fixed to sliding. Now we have 3 fixed contacts and one sliding. The left foot on the slope shuffles back and forward. Figures~\ref{shufflingExp}~and~\ref{FootShuffling} show the shuffling experiment and the normal force tracking for the end effectors respectively.

\begin{figure}[t!]
	\centering
	\includegraphics[width=0.95\columnwidth]{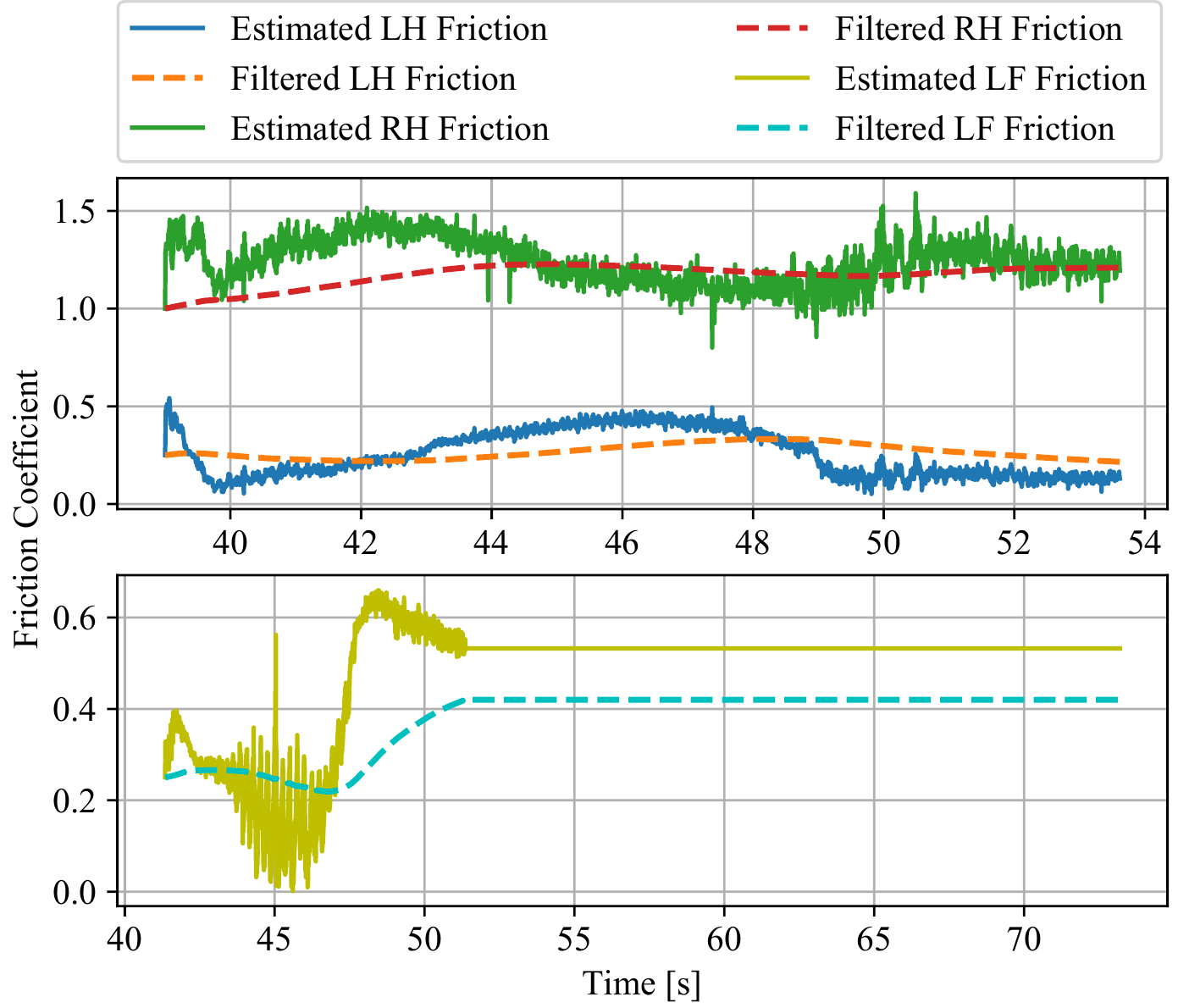} 
	\caption{Estimated and filtered dynamic friction coefficient for co-wiping (up) and shuffling scenarios (bottom).}
	\label{frictionEstimation}
\end{figure}

In the last part of the scenario HRP-4 releases both hands at a time and steps back to the floor (initial pose). During all sliding motions, online friction estimation~\eqref{filter} is executed. The estimation process begins with an initial guesses, as shown in Fig.~\ref{frictionEstimation}, and computes friction only when normal forces are above a given threshold. 

\begin{figure}[tb!]
	\centering
	\includegraphics[width=0.95\columnwidth]{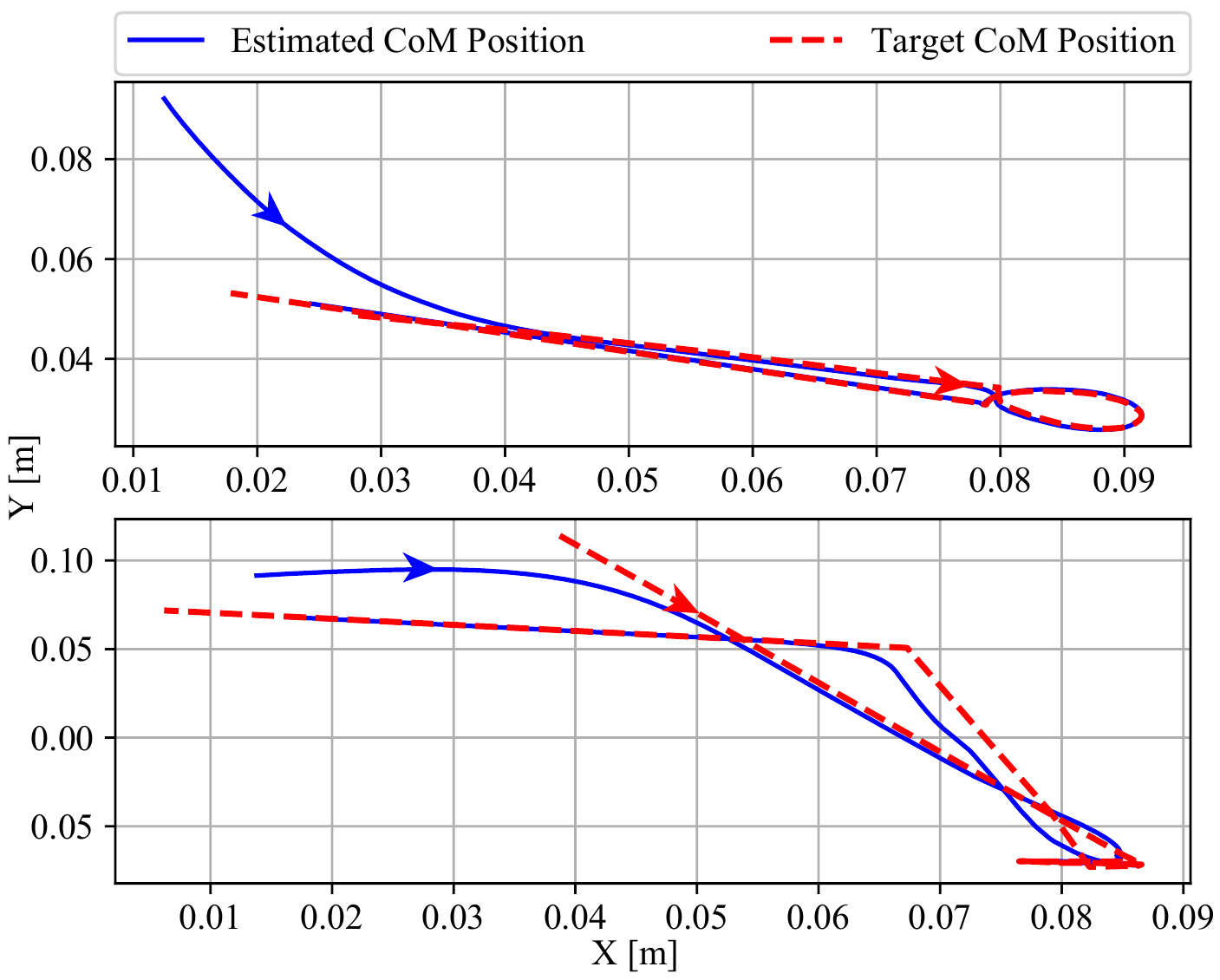}
	\caption{CoM trajectory tracking in $x$ and $y$ directions for co-wiping (up) and shuffling scenarios (bottom)}
	\label{CoMTracking}
\end{figure} 

The experiments computations were made by a laptop computer having Intel(R) Xeon(R) E-2276M CPU at $2.80$~GHz$\times12$. With this setup, the average computation times for the Chebyshev QP and the whole framework were noted as $0.2$ and $1$~ms respectively. These values are largely within the implemented real-time control loop with {\tt mc\_rtc} framework\footnotemark that is $5$~ms.
\footnotetext{\url{https://github.com/jrl-umi3218/mc_rtc}}

\section{Conclusion} 
\label{Sec_Conclusion}
In this letter, we devised a whole-body humanoid non-coplanar multi-contact motion planning and control for mixed sliding and fixed contacts that can be switched at will. Our method does not require the construction of GIWC or CoM-support polytopes thanks to a single and fast optimization problem based on the Chebyshev center. This makes it suitable for closed-loop control. It also allows exploiting all the contacts (including moving/sliding ones) for dynamic balance when this is possible, and permit balance contacts to also contribute to force tasks.

We assess our approach through complex scenarios with the HRP-4 humanoid robot: four contacts are controlled in force under dynamic balance with switches between fixed and sliding contact modes at user's will. A simple online friction estimator is implemented to update the friction coefficient of the sliding contacts. A video of the experiments\footnotemark\footnotetext{\url{https://youtu.be/cFYd9oQueRE}} and the open-source code of the controller\footnotemark\footnotetext{\url{https://github.com/SaeidSamadi/Multi-sliding-Contacts}} are available online.

In future work, we aim at improving the force tracking based on the extension of~\cite{pham2020ral} that is currently limited to translation forces (i.e. non-moments). We are also considering to extend multi-contact modalities to include soft supports. That is to say, contact that combines fixed, sliding, pushing, and rolling eventually on soft supports. We then need to develop a more sophisticated friction identification models and associated filters. These modalities would then cover almost all spectrum of possible contacts encountered in the applications we are targeting. Robustness consideration, namely w.r.t. uncertain dynamic parameters shall also be accounted for (e.g. considering relative forces instead of absolute ones). Finally, we shall cover multi-contact planning that considers such multi-modal contacts in order to achieve tasks that require accessing narrow or cumbersome passages. These skills will be deployed in real use-case scenarios defined by two large-scale manufacturing industrial partners, one of which is the continuation of~\cite{kheddar2019ram} and in physical human-robot assistive robotics~\cite{bolotnikova2020ral}.
 
\section{Acknowledgement} \label{Acknowledgement}
The authors would like to thank Dr. Pierre Gergondet for his technical supports on the {\tt mc\_rtc} framework.

\bibliographystyle{ieeetr}
\bibliography{refs}

\end{document}